\PassOptionsToPackage{unicode=true}{hyperref} 
\PassOptionsToPackage{hyphens}{url}
\pdfoutput=1
\documentclass[]{article}
\usepackage{lmodern}
\usepackage{amssymb,amsmath}
\usepackage{ifxetex,ifluatex}
\usepackage{cite}
\usepackage{fixltx2e} 
\ifnum 0\ifxetex 1\fi\ifluatex 1\fi=0 
  \usepackage[T1]{fontenc}
  \usepackage[utf8]{inputenc}
  \usepackage{textcomp} 
\else 
  \usepackage{unicode-math}
  \defaultfontfeatures{Ligatures=TeX,Scale=MatchLowercase}
\fi
\IfFileExists{upquote.sty}{\usepackage{upquote}}{}
\IfFileExists{microtype.sty}{%
\usepackage[]{microtype}
\UseMicrotypeSet[protrusion]{basicmath} 
}{}
\IfFileExists{parskip.sty}{%
\usepackage{parskip}
}{
\setlength{\parindent}{0pt}
\setlength{\parskip}{6pt plus 2pt minus 1pt}
}
\usepackage{hyperref}
\hypersetup{
            pdfborder={0 0 0},
            breaklinks=true}
\usepackage{longtable,booktabs}
\IfFileExists{footnote.sty}{\usepackage{footnote}\makesavenoteenv{longtable}}{}
\usepackage{graphicx,grffile}
\makeatletter
\def\maxwidth{\ifdim\Gin@nat@width>\linewidth\linewidth\else\Gin@nat@width\fi}
\def\maxheight{\ifdim\Gin@nat@height>\textheight\textheight\else\Gin@nat@height\fi}
\makeatother
\setkeys{Gin}{width=\maxwidth,height=\maxheight,keepaspectratio}
\ifx\paragraph\undefined\else
\let\oldparagraph\paragraph
\renewcommand{\paragraph}[1]{\oldparagraph{#1}\mbox{}}
\fi
\ifx\subparagraph\undefined\else
\let\oldsubparagraph\subparagraph
\renewcommand{\subparagraph}[1]{\oldsubparagraph{#1}\mbox{}}
\fi

\makeatletter
\def\fps@figure{htbp}
\makeatother

\date{}

\begin{document}

\hypertarget{simplellm4ad-an-end-to-end-vision-language-model-with-graph-visual-question-answering-for-autonomous-driving}{%
\section{SimpleLLM4AD: An End-to-End Vision-Language Model with Graph
Visual Question Answering for Autonomous
Driving}\label{simplellm4ad-an-end-to-end-vision-language-model-with-graph-visual-question-answering-for-autonomous-driving}}

Peiru Zheng, Yun Zhao, Zhan Gong, Hong Zhu, Shaohua Wu

IEIT Systems

\hypertarget{abstract}{%
\subsection{Abstract}\label{abstract}}

Many fields could benefit from the rapid development of the large
language models (LLMs). The end-to-end autonomous driving (e2eAD) is one
of the typically fields facing new opportunities as the LLMs have
supported more and more modalities. Here, by utilizing vision-language
model (VLM), we proposed an e2eAD method called SimpleLLM4AD. In our
method, the e2eAD task are divided into four stages, which are
perception, prediction, planning, and behavior. Each stage consists of
several visual question answering (VQA) pairs and VQA pairs interconnect
with each other constructing a graph called Graph VQA (GVQA). By
reasoning each VQA pair in the GVQA through VLM stage by stage, our
method could achieve e2e driving with language. In our method, vision
transformers (ViT) models are employed to process nuScenes visual data,
while VLM are utilized to interpret and reason about the information
extracted from the visual inputs. In the perception stage, the system
identifies and classifies objects from the driving environment. The
prediction stage involves forecasting the potential movements of these
objects. The planning stage utilizes the gathered information to develop
a driving strategy, ensuring the safety and efficiency of the autonomous
vehicle. Finally, the behavior stage translates the planned actions into
executable commands for the vehicle. Our experiments demonstrate that
SimpleLLM4AD achieves competitive performance in complex driving
scenarios.

\hypertarget{introduction}{%
\subsection{1. Introduction}\label{introduction}}

Autonomous driving has garnered significant attention from both academia
and industry over the past decade.\cite{RN6,RN7,RN8} The promise of
safer roads, reduced traffic congestion, and increased mobility for all
populations has driven rapid advancements in this field. Traditional
approaches to autonomous driving often rely on a modular pipeline
consisting of perception, prediction, planning, and control. However,
these methods can suffer from compounding errors across modules, leading
to suboptimal performance in complex and dynamic driving environments.

With the advent of large language models (LLMs),\cite{RN9,RN10} a
new opportunity arises to redefine the approach to autonomous driving.
LLMs, particularly when integrated with vision-language models
(VLMs),\cite{RN11,RN12} have shown remarkable capabilities in
understanding and generating human-like text based on visual inputs.
These capabilities can be harnessed to create more coherent and
integrated autonomous driving systems, capable of nuanced reasoning and
decision-making.

In this work, we introduce SimpleLLM4AD, an end-to-end autonomous
driving (e2eAD) method that leverages the power of VLMs. Our method
reimagines the traditional autonomous driving pipeline by structuring
the task into four interconnected stages: perception, prediction,
planning, and behavior. Each stage is framed as a series of visual
question answering (VQA) pairs, which are interlinked to form a Graph
VQA (GVQA). This graph-based structure allows the system to reason about
each VQA pair systematically, ensuring a coherent flow of information
and decision-making from perception to action.

The perception stage in SimpleLLM4AD utilizes ViT models to process raw
visual data, extracting meaningful features and identifying objects in
the driving environment. These visual insights are then translated into
a format that the language model can interpret, allowing for a more
sophisticated analysis of the scene. In the prediction stage, the system
forecasts the future states of the identified objects, considering their
potential movements and interactions. This predictive capability is
crucial for anticipating potential hazards and planning safe maneuvers.
The planning stage involves synthesizing the information gathered from
the previous stages to develop a driving strategy. This strategy is
designed to optimize safety and efficiency, taking into account the
dynamic nature of the driving environment. Finally, the behavior stage
converts the planned actions into executable commands for the vehicle,
ensuring precise control and responsiveness.

Our experiments demonstrate that SimpleLLM4AD not only achieves
competitive performance in driving benchmarks but also exhibits enhanced
robustness in complex scenarios. The integration of VLMs enables the
system to make context-aware decisions, significantly improving its
reliability and safety. Our key contributions are summarized as follows:

• We leverage the logical dependency of GVQA by utilizing the answers to
associated questions as contextual information for the current question.
This approach has been shown to significantly enhance the capabilities
of LLMs in terms of accuracy (ACC) and language score.

• We refine the prompts to further boost the performance of LLMs. We
optimize the simple question-and-answer (Q+A) format of the contexts to
simplify the contextual information, making it easier and more efficient
for LLMs to leverage previous knowledge.

• Finally, we introduce object detection branches into the LLM
optimization process, which include object localization, color
identification, and categorization. These additional branches provide
LLMs with a richer set of contextual cues, further enhancing their
performance.

\hypertarget{related-work}{%
\subsection{2. Related work}\label{related-work}}

\hypertarget{large-language-models}{%
\subsubsection{Large Language Models}\label{large-language-models}}

LLMs have rapidly evolved in recent years, demonstrating unprecedented
capabilities in natural language understanding and generation. These
models, exemplified by OpenAI's GPT-3\cite{RN13} and
GPT-4\cite{RN14}, and other open-sourced transformer-based
architectures such as the LLaMA series,\cite{RN9,RN10}
Vicuna,\cite{RN15} Baichuan,\cite{RN16}
Qwen,\cite{RN17} Yuan 2.0,\cite{RN18} Yuan
2.0-M32,\cite{RN19} leverage vast amounts of data and
sophisticated training techniques to produce human-like text and
comprehend complex linguistic structures.

One of the significant breakthroughs with LLMs has been their ability to
perform a wide range of tasks with minimal fine-tuning. This versatility
arises from the models' pre-training on diverse corpora, enabling them
to capture extensive world knowledge and linguistic patterns. The
emergence of LLMs has led to remarkable advancements in various fields,
including natural language processing (NLP), question answering, machine
translation, and text summarization.

\hypertarget{vision-language-models}{%
\subsubsection{Vision-language Models}\label{vision-language-models}}

VLMs represent a significant advancement in the field of artificial
intelligence, aiming to bridge the gap between visual and textual data.
The architecture of VLMs typically involves two main components: a
visual encoder and a language decoder. The visual encoder processes the
image to extract features, which are then combined with text
representations generated by the language decoder. This integrated
approach enables the model to perform complex reasoning tasks that
require both visual and linguistic understanding. These models combine
the strengths of vision models, such as convolutional neural networks
(CNNs)\cite{RN20,RN21} and ViT\cite{RN22,RN23,RN24,RN25}, with the
capabilities of language models, allowing for more comprehensive
understanding and interaction with multimodal inputs.

Efforts in VLMs focused on tasks like image captioning and VQA. Image
captioning involves generating descriptive textual annotations for
images, while VQA requires the model to answer questions about the
content of an image. Nowadays, more and more VLMs have demonstrated
impressive capabilities in these areas by effectively integrating visual
and textual information. Flamingo\cite{RN26} utilizes visual
and language inputs as prompts, demonstrating exceptional few-shot
performance in visual question answering. GPT-4,\cite{RN14} the
LLaVA series\cite{RN11,RN12} introduces visual instruction tuning
to enhance the instruction-following capabilities of VLMs. Concurrently,
models like KOSMOS-2\cite{RN27}, and
Qwen-VL\cite{RN28} have advanced VLMs by incorporating visual
grounding capabilities, facilitating tasks such as region description
and localization. Furthermore, PaLM-E\cite{RN29} and
EmbodiedGPT\cite{RN30} represent significant strides in
adapting VLMs for embodied applications, vastly expanding their
potential use cases. These advancements illustrate the rapid progress in
the development and application of VLMs, showcasing their ability to
handle increasingly complex multimodal tasks. As the integration of
vision and language continues to evolve, VLMs are poised to play a
pivotal role in various AI-driven domains, including autonomous driving,
healthcare, and interactive robotics.

\hypertarget{language-grounded-driving}{%
\subsubsection{Language-grounded
Driving}\label{language-grounded-driving}}

Language-grounded driving is an emerging research area that combines the
principles of NLP with AD technologies. This interdisciplinary approach
leverages the understanding and generation of natural language to inform
and enhance the decision-making processes of autonomous vehicles. By
grounding driving actions in language, these systems can achieve greater
interpretability, flexibility, and user interaction capabilities.

Recent advancements in LLMs and VLMs have propelled the field of
language-grounded driving. Models like GPTDriver,\cite{RN31}
LLM-Driver,\cite{RN32} and LMDrive\cite{RN33} offer
new possibilities for integrating language grounding into autonomous
driving. These models enable more nuanced and context-aware
interpretations of driving environments and instructions.

Language-grounded driving represents a promising frontier in autonomous
vehicle technology. By integrating natural language understanding with
driving systems, researchers are developing more interactive,
interpretable, and adaptable autonomous vehicles. Our SimpleLLM4AD
method builds upon these advancements, utilizing language grounding to
enhance each stage of the e2eAD pipeline, from perception and prediction
to planning and behavior. This integration not only improves the
vehicle's decision-making capabilities but also facilitates better
human-vehicle interaction and scenario-based training.

\hypertarget{method}{%
\subsection{3. Method}\label{method}}

\hypertarget{overall-architecture.}{%
\subsubsection{3.1. Overall architecture.}\label{overall-architecture.}}

The pipeline of our method is outlined in Figure 1. The overall
architecture comprises two main modules: a vision encoder that processes
images, and an LLM decoder that handles questions.

\textbf{Vision Encoder.} We choose InternViT-6B as the vision encoder.
InternViT-6B, a vision transformer with 6 billion parameters, was first
introduced by Chen et al.\cite{RN5} It was pre-trained using
web-scale image-text data from various sources to align with large
language models. The Query Model serves as a bridge between the vision
encoder and the LLM decoder, aligning the vision and text modalities.
This vision-text alignment component is initialized with a
multilingual-enhanced LLaMA.\cite{RN34}

\textbf{LLM Decoder.} We employ Vicuna-13B as the LLM decoder.
Vicuna-13B is an open-source LLM that was fine-tuned from LLaMA using
user-shared conversations collected from ShareGPT.31 Although different
questions share the same LLM decoder model, we have devised a GVQA
strategy to enhance the language model's capabilities and crafted
tailored prompts based on different problem types.

In our method, each key frame in the nuScenes dataset undergoes a series
of QA pairs. Initially, the six images of the nuScenes key frame are
encoded into feature maps using InternViT-6B. Next, in the Query module,
the image features interact with 96 learnable queries and the question
text from the QA pair. The Query module transforms the image tokens
generated by InternViT-6B into representations that are aligned with the
LLMs. Finally, the output from the Query module is fed into Vicuna-13B
to generate an answer. Notably, the answer generated in the previous
step is combined with the subsequent question to form a context-rich
question. By iterating this process, the pipeline progresses round by
round to accomplish the end-to-end autonomous driving (e2eAD) tasks step
by step.

This modular approach guarantees that each processing stage---from
visual encoding to vision-text alignment and ultimately language
generation---is optimized for its distinct function while maintaining
seamless integration throughout the entire pipeline. This architecture
not only elevates the system's capability to handle intricate visual and
linguistic inputs but also ensures a coherent flow of information,
enabling more precise and context-aware decision-making in autonomous
driving scenarios.

\includegraphics{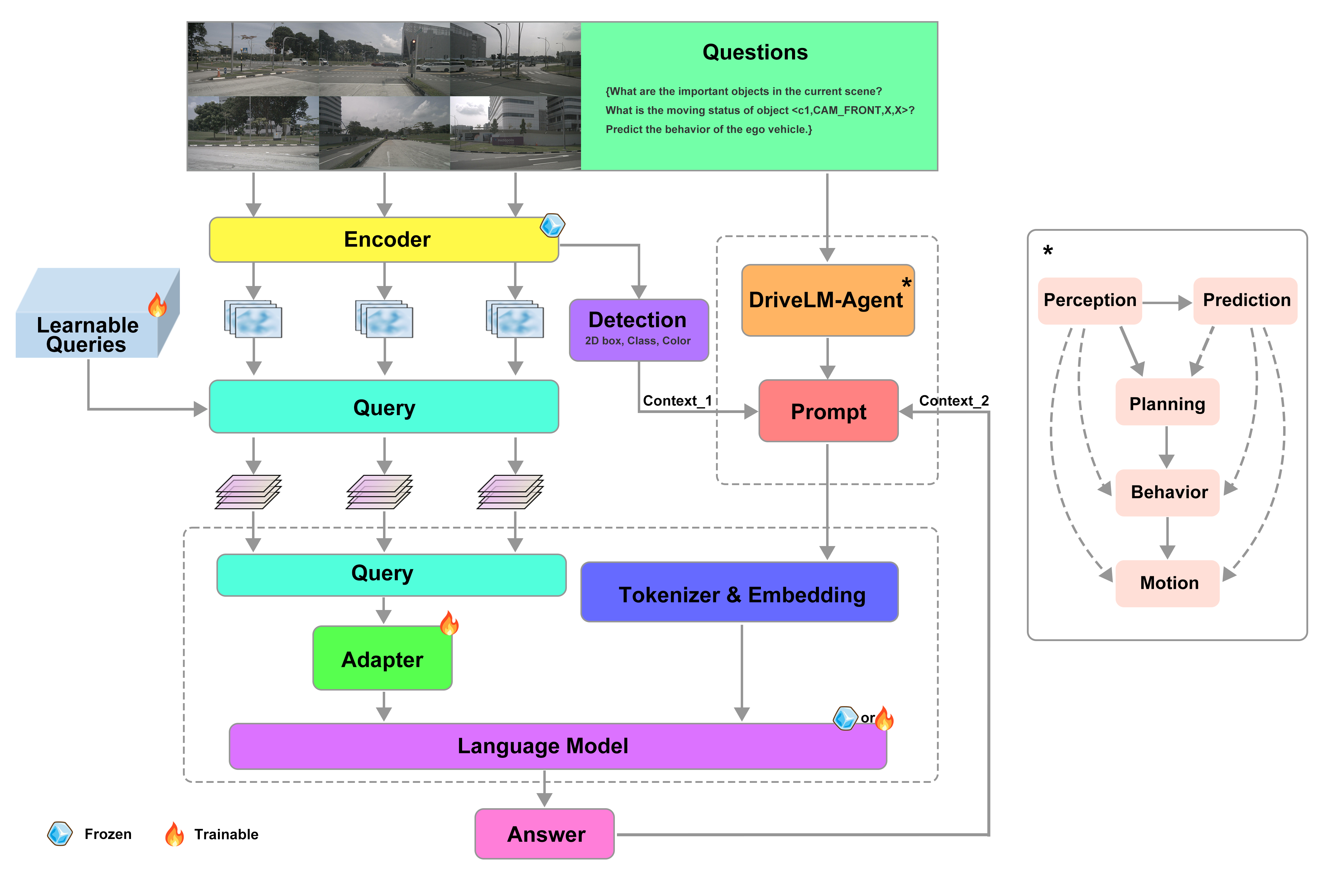}

Figure 1. Overall architecture.

\hypertarget{gvqa-logical-dependency}{%
\subsubsection{3.2. GVQA logical
dependency}\label{gvqa-logical-dependency}}

The SimpleLLM4AD method encompasses a sequence of four stages, each
intricately linked by the logical dependencies of the QA pairs they
contain. As depicted in Fig. 2, the logical dependency of GVQA is
graphically represented, illustrating the interconnected nature of the
QA pairs throughout the process. The diagram consists of two primary
elements: nodes (N) and edges (E). Nodes (N) symbolize individual QA
pairs, and the edges (E) that connect them denote the logical
dependencies that exist between them. Specifically, the answer from a
preceding node (N\textsubscript{P}) serves as contextual information for
the subsequent node (N\textsubscript{S}).

In the context of AD, understanding this logical dependency is crucial
for the system's decision-making process. For instance, during the
Perception stage, the system must identify key objects in the current
scene that will be considered for future reasoning and driving
decisions. This initial identification sets the stage for the Prediction
stage, where the system assesses the motion status of objects and
predicts their potential future states.

The Planning stage then utilizes the information gathered during
Perception and Prediction to formulate a safe and effective course of
action for the ego vehicle. This involves considering the potential
actions that could be taken in response to the identified objects and
predicting the outcomes of those actions, including the likelihood of
collisions and the safety of the proposed maneuvers.

Figure 2 elucidates the logical flow from one stage to the next, with
each node representing a critical decision point or informational
milestone. For example, (c1,CAM\_FRONT,714.3,503.6) represents the
identification and initial assessment of an object captured by the front
camera. The subsequent nodes then build upon this information, asking
questions about the object's movement status and how it might interact
with other objects in the scene, such as node
(c3,CAM\_FRONT,1300.8,531.7), which could represent another vehicle or
significant obstacle.

The edges connecting these nodes indicate the flow of logic and the
progression of the system's thought process. For instance, the question
of whether object `c1' will be in the moving direction of object `c3' is
predicated on the answers derived from the previous nodes. Similarly,
the decision-making process regarding the ego vehicle's actions is
contingent upon the predictions and assessments made in the preceding
stages.

By mapping out these logical dependencies, the SimpleLLM4AD method
ensures a coherent and systematic approach to autonomous driving
decision-making. This structured methodology not only aids in the
development of more sophisticated AI systems but also enhances the
transparency and reliability of the decisions made by these systems.

\includegraphics{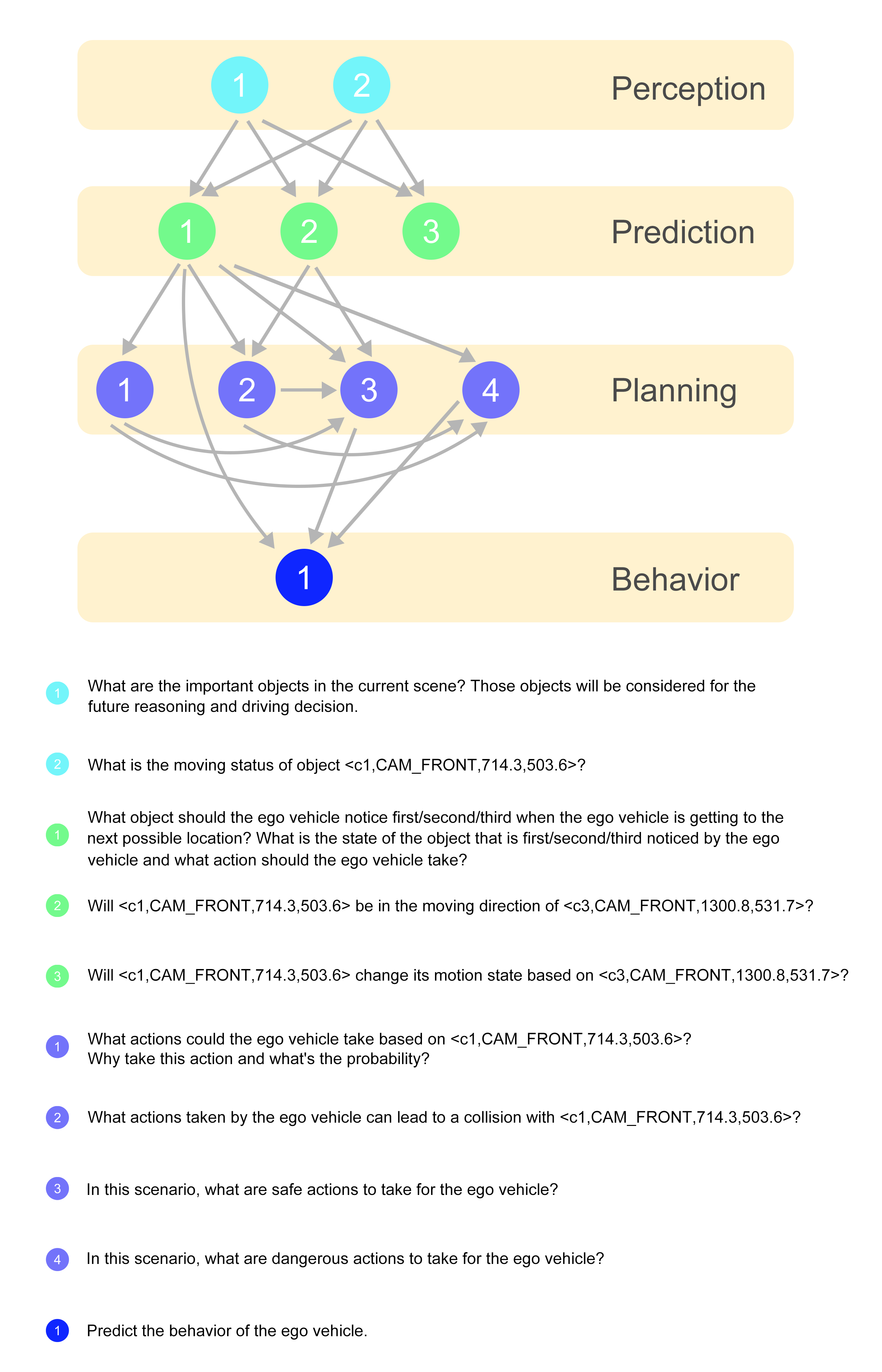}

Figure 2. GVQA logical dependency.

\hypertarget{experiments}{%
\subsection{4. Experiments}\label{experiments}}

\hypertarget{datasets-and-metrics}{%
\subsubsection{4.1. Datasets and Metrics}\label{datasets-and-metrics}}

\hypertarget{dataset}{%
\paragraph{Dataset}\label{dataset}}

The datasets used for fine-tuning and evaluation is
DriveLM-nuScenes.\cite{RN4} DriveLM-nuScenes is a
comprehensive dataset specifically designed for developing and
benchmarking autonomous driving models. It consists of a training set of
4072 frames and a validation set of 799 frames, providing a robust
foundation for training and testing. DriveLM-nuScenes incorporates both
scene-level descriptions and frame-level QA pairs, divided into three
categories: perception, prediction, and planning. This structure ensures
a comprehensive understanding of the driving scenarios by addressing
various aspects of the driving process.

Perception: This category involves questions related to the thorough
examination of the entire frame. While some questions are manually
annotated, prompts are designed to generate questions based on the
observational facets of objects within the scene, leveraging ground
truth from nuScenes and OpenLane-V2 datasets.

Prediction: This category includes inquiries regarding the future states
of key objects and the ego vehicle in the current frame, focusing on the
reasoning process behind these predictions. Due to the complexity of
these predictions, the answers are manually annotated.

Planning: This involves questions related to the subsequent actions of
the ego vehicle within the current frame. Similar to prediction,
planning questions are challenging and require manual annotation for the
reasoning process.

Each key object within the QA pairs is encoded as c tags in the format
\textless{}c, CAM, x, y\textgreater{}, where c is the identifier, CAM
indicates the camera, and x, y are the coordinates of the object's 2D
bounding box in the respective camera's coordinate system. The dataset
also includes a dictionary in each key frame, recording basic
information about key objects such as bounding box size, category,
moving state, and visual description.

\hypertarget{metrics}{%
\paragraph{Metrics}\label{metrics}}

To evaluate the performance of our models, we use a set of
well-established metrics tailored to different tasks within the
autonomous driving domain. The metrics are divided into three main
categories: P1-3 VQA Metrics, Behavior Task Metrics, and Motion Task
Metrics.

\textbf{P1-3 VQA Metrics}: These metrics assess the performance of
perception, prediction, and planning question-answering (QA) tasks.
Common Visual Question Answering (VQA) metrics are employed, alongside a
new GPT score for a more nuanced semantic evaluation.

\begin{quote}
\textbf{BLEU}: Measures n-gram overlap between generated text and
reference text, focusing on precision but often insensitive to semantic
nuances.

\textbf{ROUGE\_L}: Focuses on recall by calculating the longest common
subsequence between model outputs and reference answers.

\textbf{METEOR}: Considers precision, recall, stemming, synonymy, and
word order, offering a detailed alignment between model outputs and
references.

\textbf{CIDEr}: Utilizes a combination of BLEU and vector space models,
calculating semantic consistency through cosine similarity of n-gram
TF-IDF vectors.

\textbf{SPICE}: Parses text into a syntactic dependency tree and maps it
to a scene graph, computing F-scores to evaluate the alignment between
predicted and reference scene graphs.

\textbf{GPT Score}: Provided by ChatGPT, this metric assesses prediction
quality through a more rational score based on robust reasoning
capabilities, assigning a score between 0 and 100.
\end{quote}

\textbf{Behavior Task Metrics}: These metrics evaluate behavior
predictions, focusing on classification accuracy of behavior, behavior
speed, and behavior steer.

\begin{quote}
\textbf{Classification Accuracy}: Evaluates the accuracy of behavior
predictions by comparing the predicted behaviors of the ego vehicle to
the ground truth. The ground truth trajectory points are categorized
into speed and steering behaviors, and the accuracy is calculated based
on the agreement between predicted and actual behaviors.
\end{quote}

These metrics ensure a comprehensive evaluation of the model's
performance across various aspects of autonomous driving, from
perception and prediction to planning and behavior prediction.

\hypertarget{implementation-details}{%
\subsubsection{4.2. Implementation
Details}\label{implementation-details}}

In our method, the SimpleLLM4AD model is finetuned with DriveLM-nuScenes
dataset. We use the pretrained weights of InternViT-6B and keep it
frozen. QLLaMA and 96 queries are trainable during finetuing. The
pretrained Vicuna-13B could be totally frozen or tuned by
parameter-efficient fine-tuning (PEFT)\cite{RN36} methods such
as LoRA\cite{RN37}. The image resolution is set to 224 × 224.
We finetuned the model on NVIDIA GPU with a learning rate of 1e-4 and a
global batch size of 16.

\hypertarget{test-results-on-drivelm-nuscenes}{%
\subsubsection{4.3. Test Results on
DriveLM-nuScenes}\label{test-results-on-drivelm-nuscenes}}

All models in Table 1 are fine-tuned on the training split of the
DriveLM-nuScenes dataset and tested on the test set of the same dataset.
The DriveLM baseline,\cite{RN4} which fine-tunes
LLaMA-Adapter-V2\cite{RN39}, achieves a final score of 32.4,
primarily due to its poor performance on the accuracy (acc.) metric. Cho
et al. fine-tune LLaVA-1.5\cite{RN11} and
Cube-LLM,\cite{RN38} resulting in a significant improvement
compared to the baseline.

Our method, SimpleLLM4AD, demonstrates a marked enhancement in both the
accuracy and language scores, leading to a final score of 52.7. The
partial result presentation is illustrated in Figure 3. This substantial
improvement highlights the effectiveness of our approach. The detailed
methodology and analysis of this result will be expounded in the
following section.

\includegraphics[width=5.64329in,height=3.0287in]{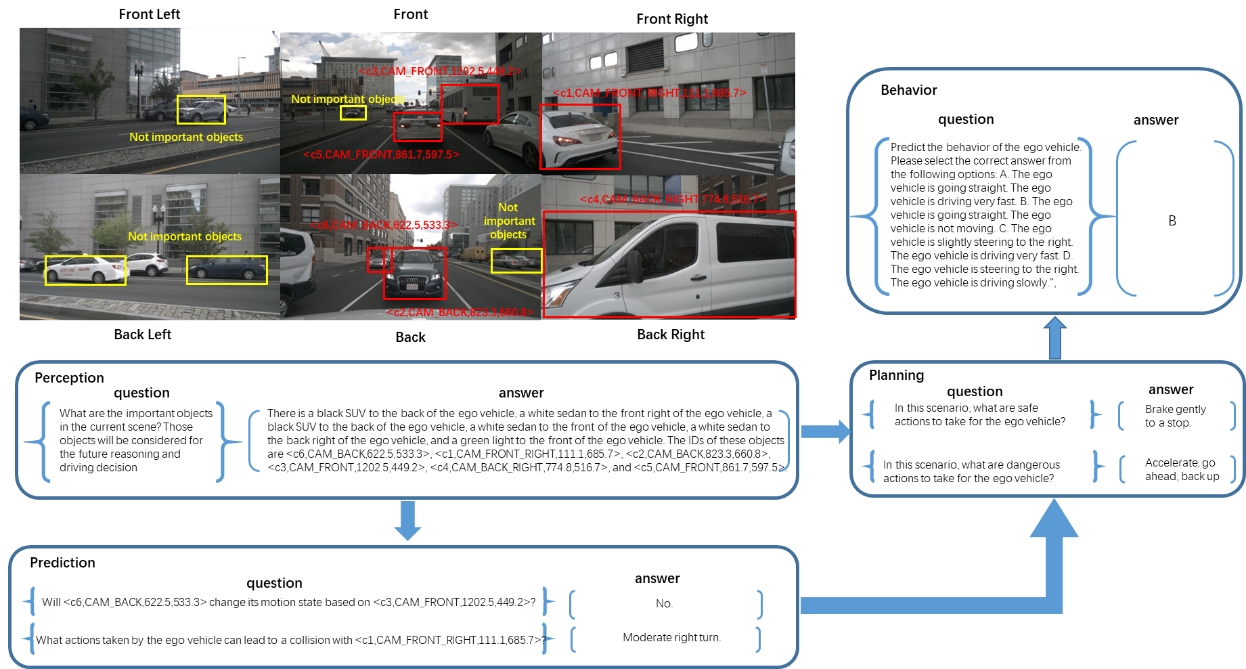}

Figure 3. \textbf{Result Presentation}.

Table 1. DriveLM-nuScenes Benchmark. (Test set)

\begin{longtable}[]{@{}llllllll@{}}
\toprule
Method & \emph{Acc.} & \emph{ChatGPT} & \emph{Match} &
\emph{BLEU\textsubscript{1}} & \emph{ROUGE\textsubscript{L}} &
\emph{CIDE\textsubscript{r}} & \emph{Final Score}\tabularnewline
\midrule
\endhead
DriveLM baseline\cite{RN4} & 0.0 & 65.1 & 28.3 & 5.0 & 8.4 &
9.9 & 32.4\tabularnewline
LLaVA-1.5\cite{RN11} & 38.5 & 53.5 & 26.1 & 15.8 & 14.3 & 30.3 &
36.1\tabularnewline
Cube-LLM\cite{RN38} & 38.5 & \textbf{89.4} & \textbf{39.0} &
16.3 & 20.4 & \textbf{31.3} & 50.1\tabularnewline
SimpleLLM4AD & \textbf{66.5} & 57.4 & 35.3 & \textbf{75.7} &
\textbf{73.4} & 15.3 & \textbf{52.7}\tabularnewline
\bottomrule
\end{longtable}

\hypertarget{ablation}{%
\subsubsection{4.4. Ablation}\label{ablation}}

During our exploration, we conducted training and inference of
SimpleLLM4AD under various settings. The differences among these
solutions primarily involve the treatment of prompts and the methods for
detecting key objects. The baseline performance shown in Table 1 was
obtained by using LLaMA-Adapter-V2 for evaluation on a validation set
that we created. This validation set was extracted from the original
training set by selecting one scene out of every six.

Table 2. DriveLM-nuScenes results of different settings. (Validation set
split by us)

\begin{longtable}[]{@{}llllll@{}}
\toprule
Method & \emph{Acc.} & \emph{Match} & \emph{BLEU\textsubscript{1}} &
\emph{ROUGE\textsubscript{L}} &
\emph{CIDE\textsubscript{r}}\tabularnewline
\midrule
\endhead
Baseline & 60.1 & 1.16 & 57.6 & 65.9 & 3.3\tabularnewline
Version A & 62.7 & 2.94 & 66.8 & 68.4 & 14.1\tabularnewline
Version B & 63.4 & 2.94 & 72.2 & 72.4 & 17.0\tabularnewline
Version C & \textbf{67.7} & 3.43 & 74.8 & 71.9 & 15.0\tabularnewline
Version D & 67.5 & 3.55 & \textbf{80.5} & \textbf{75.1} &
\textbf{33.2}\tabularnewline
Version E & - & \textbf{24.6} & \textbf{-} & \textbf{-} &
\textbf{-}\tabularnewline
\bottomrule
\end{longtable}

Notice: In Table 2, the "match" only measures the box coordinate
matching; unlike in Table 1, where the "match" includes both box
coordinate matching and the ChatGPT score.

\hypertarget{chain-of-thought}{%
\paragraph{Chain of Thought}\label{chain-of-thought}}

Chain-of-Thought (CoT) is an approach for prompting, in which one
includes the intermediate steps of reasoning within the prompt
(intermediate ``thoughts''), besidesthe task
input/output.\cite{RN40} CoT was shown to significantly improve
the capability of LLMs to solve problems without resorting to any model
updates.

In SimpleLLM4AD, we employ CoT by use the answer of N\textsubscript{P}
as contextual information for N\textsubscript{S}, and N\textsubscript{S}
would always be the next QA pair of N\textsubscript{P} in the dataset.
Comparing to the DriveLM baseline, Version A that utilizing CoT show a
remarkable improvement in acc. And language score.

In the DriveLM-nuScenes dataset, `What are the important objects in the
current scene?' question would always be the first question in each
frame, which we call it N\textsubscript{0}. In Version B, we apply
N\textsubscript{0} + N\textsubscript{P} as context for each
N\textsubscript{S}.

\hypertarget{graph-of-thought}{%
\paragraph{Graph of Thought}\label{graph-of-thought}}

While CoT only support unidirectional single branch of thought when
prompting LLMs step by step, graph of thought (GoT) enable current task
use any previous QA pairs as context and would allow any subsequent
questions refer to the results of current task. After trying different
logical dependency graphs, the prominent one is shown in Fig. 2. By
using GoT to arrange context like this, Version C get a prominent
improvement.

\hypertarget{prompt-optimization}{%
\paragraph{Prompt Optimization}\label{prompt-optimization}}

Refined prompt would contribute to the performance of LLMs. In the
solution of DriveLM baseline, the context that pass to the subsequent
question would be previous Q + A, making the context redundant and hard
for LLMs to understand. Therefore, we optimize the simple Q + A format
of the contexts to simplify the context, making it easier and more
efficient for LLMs to use the previous information.

The original answer of N\textsubscript{0} is:

`There is a red car to the front of the ego vehicle, a white SUV to the
front of the ego vehicle, a white sedan to the front of the ego vehicle,
a black sedan to the front of the ego vehicle, and a red light to the
front of the ego vehicle. The IDs of these objects are
\textless{}c1,CAM\_FRONT,714.3,503.6\textgreater{},
\textless{}c2,CAM\_FRONT,993.3,503.3\textgreater{},
\textless{}c3,CAM\_FRONT,1300.8,531.7\textgreater{},
\textless{}c4,CAM\_FRONT,892.5,507.5\textgreater{}, and
\textless{}c5,CAM\_FRONT,712.6,361.8\textgreater{}.'

As you can see, the color/class information and the coordinate
information of the same object are separately described in two sentence,
which could make LLMs confused. Therefore, when using it as the context
for subsequent question, we format the answer as:

`\textless{}c1,CAM\_FRONT,714.3,503.6\textgreater{} is a red car to the
front of the ego vehicle.'

We only give the information of the object mentioned in the current
question instead of all objects detected in N\textsubscript{0}.

Besides, we modify other QA pairs into declarative sentence when used as
a context. For example, QA pair like `Would
\textless{}c1,CAM\_FRONT,714.3,503.6\textgreater{} be in the moving
direction of the ego vehicle? No.' would be modified to
`\textless{}c1,CAM\_FRONT,714.3,503.6\textgreater{} would not be in the
moving direction of the ego vehicle.' when used as a context.

By reformatting the QA pairs, brief and informative context provide
improvement of the language score in Version D.

\hypertarget{detection-of-the-key-objects}{%
\paragraph{Detection of the Key
Objects}\label{detection-of-the-key-objects}}

Format\_instruction is "Input six images in turn. The first of six
images is \textless{}CAM\_FRONT\textgreater{}, which is in the front of
the ego vehicle. The second of six images is
\textless{}CAM\_FRONT\_LEFT\textgreater{}, which is in the front left of
the ego vehicle. The third of six images is
\textless{}CAM\_FRONT\_RIGHT\textgreater{}, which is in the front right
of the ego vehicle. The fourth of six images is
\textless{}CAM\_BACK\textgreater{}, which is in the back of the ego
vehicle. The fifth of six images is
\textless{}CAM\_BACK\_LEFT\textgreater{}, which is in the back left of
the ego vehicle. The sixth of six images is
\textless{}CAM\_BACK\_RIGHT\textgreater{}, which is in the back right
left of the ego vehicle. \textless{}number, number\textgreater{} is
object box center coordinate in the image (1600*900)."

To enhance the performance, we have intergrated Format\_instruction to
guide LLM. To accurately detect and classify objects, we leverage
existing, well-established detection networks, such as dinov2, which
provide us with robust target detection output. In addition, we have
trained a specialized detection classification network that goes beyond
mere object detection. This network is fine-tuned to identify key
attributes of the detected objects, including their color, precise
position, and even orientation. This approach allows us to generate
detailed descriptions such as "These is a red car to the front of the
ego vehicle, and the box center is {[}714.3,503.6{]}."

By incorporating this rich information about the location, color, and
class of objects within the ego vehicle's environment, we have achieved
a significant improvement in the match score of Version E. Especially,
the positioning accuracy of the box center has significantly improved to
24.6\%.

\hypertarget{conclusion}{%
\subsection{5. Conclusion}\label{conclusion}}

In this paper, we present SimpleLLM4AD, a multi-modal language model
that achieves competitive performance in complex driving scenarios. The
integration of VLM allows for more nuanced and context-aware
decision-making, significantly improving the robustness and reliability
of the autonomous driving system. This approach also highlights the
potential for LLMs to enhance multimodal AI applications, paving the way
for further advancements in the field of e2eAD. In the future, we will
continue to optimize box positioning as well as the alignment of
multi-frame vision-language large models and other related directions.

\bibliographystyle{unsrt}

\cite{RN6}
\cite{RN7}
\cite{RN8}
\cite{RN9}
\cite{RN10}
\cite{RN11}
\cite{RN12}
\cite{RN13}
\cite{RN14}
\cite{RN15}
\cite{RN16}
\cite{RN17}
\cite{RN18}
\cite{RN19}
\cite{RN20}
\cite{RN21}
\cite{RN22}
\cite{RN23}
\cite{RN24}
\cite{RN25}
\cite{RN26}
\cite{RN27}
\cite{RN28}
\cite{RN29}
\cite{RN30}
\cite{RN31}
\cite{RN32}
\cite{RN33}
\cite{RN5}
\cite{RN34}
\cite{RN35}
\cite{RN4}
\cite{RN36}
\cite{RN37}
\cite{RN39}
\cite{RN38}
\cite{RN40}

\bibliography{ref}

\end{document}